\documentclass[sigconf,bookmarksnumbered,unicode,colorlinks,linkcolor=blue]{acmart}
\acmSubmissionID{588}

\usepackage{booktabs} 
\usepackage{tabularx}

\newcommand{\finalrev}[1]{{\color{black}{#1}}}

\usepackage[ruled]{algorithm2e} 

\SetAlFnt{\small}
\SetAlCapFnt{\small}
\SetAlCapNameFnt{\small}
\SetAlCapHSkip{0pt}

\acmJournal{TOG}

\begin{document}
\title{UltraGlove: Hand Pose Estimation with MEMS-Ultrasonic Sensors}

\settopmatter{authorsperrow=4}
\author{Qiang Zhang}
\orcid{0000-0002-4483-1039}
\affiliation{%
 \institution{Princeton University, USA} \country{}
 }
\email{qz9238@princeton.com}

\author{Yuanqiao Lin}
\orcid{0000-0003-3627-6968}
\affiliation{%
 \institution{Princeton University, USA} \country{}
 }
\email{yuanqiao@princeton.edu}

\author{Yubin Lin}
\orcid{0000-0003-4476-9383}
\affiliation{%
 \institution{Princeton University, USA} \country{}
 }
\email{yubinlin@princeton.edu}

\author{Szymon Rusinkiewicz}
\orcid{0000-0002-4253-2588}
\affiliation{%
 \institution{Princeton University, USA} \country{}
 }
\email{smr@princeton.edu}

\begin{abstract}
Hand tracking is an important aspect of human-computer interaction and has a wide range of applications in extended reality devices. However, current hand motion capture methods suffer from various limitations. For instance, visual hand pose estimation is susceptible to self-occlusion and changes in lighting conditions, while IMU-based tracking gloves experience significant drift and are not resistant to external magnetic field interference. To address these issues, we propose a novel and low-cost hand-tracking glove that utilizes several MEMS-ultrasonic sensors attached to the fingers, to measure the distance matrix among the sensors. Our lightweight deep network then reconstructs the hand pose from the distance matrix. Our experimental results demonstrate that this approach is both accurate, size-agnostic, and robust to external interference. We also show the design logic for the sensor selection, sensor configurations, circuit diagram, as well as model architecture.
\end{abstract}


%
%




\maketitle

\section{Introduction}

Hand tracking is an essential human-computer interaction (HCI) technology for a variety of applications, such as virtual training systems, virtual/augmented reality (VR/AR) systems, and robotic dexterous manipulation. For example, hand tracking is widely used in the film-making industry, as hand poses captured from actors are reprojected to animated characters, known as avatars, for better realism. Similarly, hand tracking benefits athletes, since they are able to record their motions and later examine them to refine their technique. Hand tracking also has important implications for the field of human-robot interaction. Humanoid robots with dexterous hands can adapt to complicated and dangerous scenarios and replace human labor. If captured hand poses and motions are made available, they can either be used in the teleoperation of robots or serve as training data for learned controllers.



Existing hand-tracking systems can be categorized by their sensing mechanisms: vision-based, IMU-based, and stretch-based. Visual tracking directly predicts hand motions from one or more RGB or RGB-D cameras, but such systems are sensitive to self-occlusion and limited to the field of view: they are prone to failure when hands are obscured from the cameras during manipulation. Additionally, background variations, such as insufficient lighting or excessive movements, also interfere with the extraction of hand poses, leading to low accuracy and inconsistent results. Both inertial- and stretch-based hand tracking can be used without the constraint that the hand must remain inside a camera's field of view, but they still have their drawbacks. Inertial-based measurements are taken from a grid of inertial measurement units (IMUs) attached to a glove, but those units are sensitive to external magnetic interference and lack long-term stability due to sensor drift. Additionally, they cannot distinguish between different finger poses with similar orientations, which limits their applicability in certain scenarios. Stretch-based methods rely on stretch sensors attached to the fingers to measure the degree of bending, but they cannot be easily adapted to people with different hand sizes and cannot easily differentiate between open- and closed-finger poses (i.e., distinguishing whether fingers are separated laterally).

Overall, the potential applications of hand-tracking technology are diverse and have far-reaching impacts from entertainment to manufacture. However, existing hand-tracking methods have certain limitations, such as low accuracy, low robustness to external interference, and lack of adaptability to different hand sizes. These drawbacks restrict their applications and hinder the deployment of hand tracking in high-compliance scenarios, such as robot control or remote surgery. Therefore, controllers are still widely used for hand pose commands, which require users to map an intuitive action (i.e. grabbing an object) to an abstract interaction (e.g., pressing a button on the controller).  This reduces the naturalness and available degrees of freedom of the interaction, often leading to unintuitive cognitive mapping, time-consuming training, and reduced precision. By developing a low-cost and accurate hand-tracking solution, this paper aims to make this technology more accessible and widely applicable, paving the way for further innovations in the field.

In this paper, we propose a novel and low-cost solution for hand tracking using a glove with multiple micro-electromechanical system (MEMS) ultrasonic sensors positioned throughout the hand. At run time, we measure the pairwise distances among the sensors and reconstruct the hand pose using a lightweight deep neural network. We have conducted extensive experiments to evaluate our method's performance in both mechanical hands with quantitative metrics and in human hands with qualitative metrics. The results demonstrate that our approach achieves high accuracy and is robust to interference under challenging scenarios that existing methods cannot handle, thus making it suitable for various HCI devices, virtual training systems, and robotic dexterous manipulation.

Our main contributions are as follows:
\begin{itemize}
\item We design and build a glove that integrates multiple MEMS-ultrasonic sensors, obtaining a pairwise distance matrix for each hand pose. We also develop a circuit and implement the corresponding embedded system to read the raw sensor data and measure the distance matrix in real time.
\item We propose a lightweight deep neural network model for accurate and real-time 3D hand pose estimation based on the raw sensor data returned from the sensor. We collect a training dataset and evaluate the performance of the proposed model, demonstrating its effectiveness through sim-to-real transfer learning.
\item We provide an in-depth analysis of the design philosophy for the raw sensor selection, sensor configurations, microcontroller unit (MCU), and circuit architecture. We also conduct an ablation study on the proposed model to evaluate the contribution of each component.
\end{itemize}

\section{Related Work}
\subsection{Visual Hand Pose Estimation}

There has been significant progress in hand pose estimation using RGB or RGB-D cameras. For example, marker-based data gloves have been proposed~\cite{wang2009real,han2018online}, which require colored or optical markers to be attached to the glove and rely on external cameras to estimate pose. 

In the absence of markers, some systems have proposed identifying 2D keypoints in images of hands, most recently based on convolutional neural networks that produce keypoint heatmaps~\cite{cai20203d,iqbal2018hand}. Some methods \cite{zimmermann2017learning,spurr2020weakly,mueller2018ganerated,cai2018weakly,wang2020rgb2hands} directly predict the 3D skeleton from a single image. For example, \cite{cai2018weakly} proposes a weakly-supervised 3D hand pose estimation algorithm from monocular RGB images. With the proven success of transformer network architectures, some papers \cite{lin2021end,lin2021mesh,li2022interacting} have proposed  transformer-based or attention-based networks for hand pose estimation. For example, \cite{li2022interacting} uses an attention mechanism to model both pose and shape, exploiting the MANO~\cite{romero2022embodied} prior.

To provide robustness to occlusion, some works focus on multi-view fusion via triangulation~\cite{simon2017hand},  post-inference optimization~\cite{han2020megatrack}, or latent-feature fusion~\cite{he2020epipolar,iskakov2019learnable,remelli2020lightweight}. For example, \cite{han2022umetrack} proposes a differentiable end-to-end architecture for multi-view camera fusion and temporal fusion to improve performance and robustness. Two-hand reconstruction exacerbates the occlusion problem, and many works have addressed occlusion- and collision-aware two-hand joint pose estimation, such as \cite{fan2021learning,kim2021end,moon2020interhand2,rong2021monocular,zhang2021interacting}.

RGB-D cameras provide extra information for hand tracking, relative to simple RGB images. Many papers have proposed deep-learning-based algorithms for single-hand tracking \cite{xiong2019a2j,mueller2019real,moon2018v2v,tang2015opening,oikonomidis2011efficient,tang2014latent} or two-hand tracking\cite{kyriazis2014scalable,mueller2019real,oikonomidis2012tracking,tzionas2016capturing} from a single depth image. For example, \cite{tang2015opening} shows a new hierarchical sampling optimization method to regress the full pose from a depth image via surrogate energy selection.

\subsection{IMU-based Data Gloves}
Many data gloves use Inertial Measurement Units (IMUs) for hand tracking. Researchers have explored systems based on different numbers of IMUs, ranging from  12~\cite{hu2020flexible} to 15~\cite{fang2017development}, 16~\cite{connolly2017imu,chang2019sensor}, or 18\cite{lin2018design}. There are also many works focusing on full body pose reconstruction via sparse (only 6) IMUs, \cite{von2017sparse,huang2018deep,jiang2022transformer}. The major drawback of this IMU-based solution is that the raw sensor is sensitive to external magnetic fields, which can cause measurement drift and requires calibration from time to time. Another disadvantage is that the absolute tracking accuracy is limited due to the need to integrate the output of gyroscopes and accelerometers, which is sensitive to noise and miscalibration.



\subsection{Stretch-sensor-based Data Gloves}
Gloves based on stretch sensors have been explored in the context of gesture recognition, such as \cite{ryu2018knitted,o2017language,hammond2014toward,lorussi2005strain}. However, their classification methods can not decode full continuous hand pose. In contrast, \cite{park2017soft,chossat2015wearable} propose using stretch sensors for continuous pose estimation, but there is no qualitative regression accuracy reported. \cite{glauser2019interactive} presents a promising approach for obtaining continuous pose using a stretch-sensor-based glove. However, the system can not distinguish the opening and closing state of the palm according to their website video~\cite{stretchvideo}, and the fabrication of such a glove is complicated.

\subsection{Other Sensor Data Gloves}
There are other hand-tracking solutions with different sensors. For bend-sensor data gloves (\cite{zheng2016development,shen2016soft,ciotti2016synergy}), the number of degrees of freedom is much less than that of a human hand, and increasing the number of bend sensors leads to a high complexity of glove design and may hinder dexterous movement. Trackers based on electromyography (EMG), such as \cite{liu2021neuropose}, require initialization and calibration for every new user and under different sensor locations. Also, the same hand poses with different forces may lead to completely different EMG signals, severely compromising the algorithm's accuracy. Electronic skin sensor solutions such as \cite{kim2022substrate} cannot decode the full hand pose and can only be used for some specific applications. \finalrev{\textit{Digits}~\cite{kim2012digits} proposes to estimate 3D hand pose through a wrist-worn IR camera.  However, it is sensitive to occlusion and requires the hand to be always inside the field of view, reducing the allowable space of wrist angles by half.}

Different from all the hand pose estimation methods mentioned above, we propose a novel data glove via ultrasonic sensors. Some previous works also use ultrasound sensing for \finalrev{motion capture, such as body capture: \cite{vlasic2007practical,laurijssen2015three,qi2014wearable,sato2011design}.  However, their capture quality is relatively low and cannot be replicated to the hand pose scale. There are also some ultrasonic-based hand gesture recognition works}, such as \cite{yang2018towards,yang2020wearable}. However, their methods can only solve the hand gesture classification task with limited categories and lack continuous motion decoding. However, our data glove can predict the full hand pose in a continuous way: on the low-level sensor side, these ultrasonic sensors measure their absolute distances to other sensors and return the mutual pairwise distance matrix in real time with high refresh rate and high spatial resolution. On the high-level algorithm side, our deep network takes this matrix as the input and predicts the hand pose. 

\finalrev{In summary, our ultrasonic-based data glove overcomes the self-occlusion challenges in visual tracking systems, overcomes the low robustness to external magnetic fields of IMU-based systems, and overcomes the opening-closing ambiguity of stretch-based systems, while providing high accuracy.}

\section{Glove System Design}

\subsection{MEMS-ultrasonic Sensor Introduction}
Traditional ultrasonic distance sensors use piezoelectric crystals to generate and receive high-frequency sound waves. Transducers convert electrical energy into mechanical vibrations, creating sound waves that are detected and converted back to electrical signals.  The time delay between transmission and reception is used to calculate the distance between the transmitter and receiver. Traditional sensors are typically large compared with finger size. Moreover, their distance measurement accuracy level is around $10-20$ mm and they have relatively low beam widths, both of which do not satisfy our system requirements. Please refer to Section~\ref{sec:otherultra} for more details.

In contrast, ultrasonic sensors based on micro-electromechanical systems (MEMS) are built with micromachining technology and thus are small and highly sensitive. MEMS technology allows for the creation of miniaturized, integrated sensors that can be mass-produced at low cost. Compared with piezoelectric-based sensors, they have a smaller size, lower power consumption, and most importantly, their accuracy level is much higher: we will illustrate how much accuracy they can achieve in the experiment section.

Here we choose the CH101 ultrasonic sensor from TDK, with a 4x4x2mm size. In our application scenario, we need the beam angle of the ultrasonic sensor to be as wide as possible, and the CH101 satisfies this need\,---\,in our experiments, its horizontal and vertical beam angles can be as wide as 150 degrees. This is necessary to avoid missing measurements when we attach these sensors to the fingers and measure their pairwise distance matrix.

\begin{figure}
\centering
  \includegraphics[width=0.95\columnwidth]{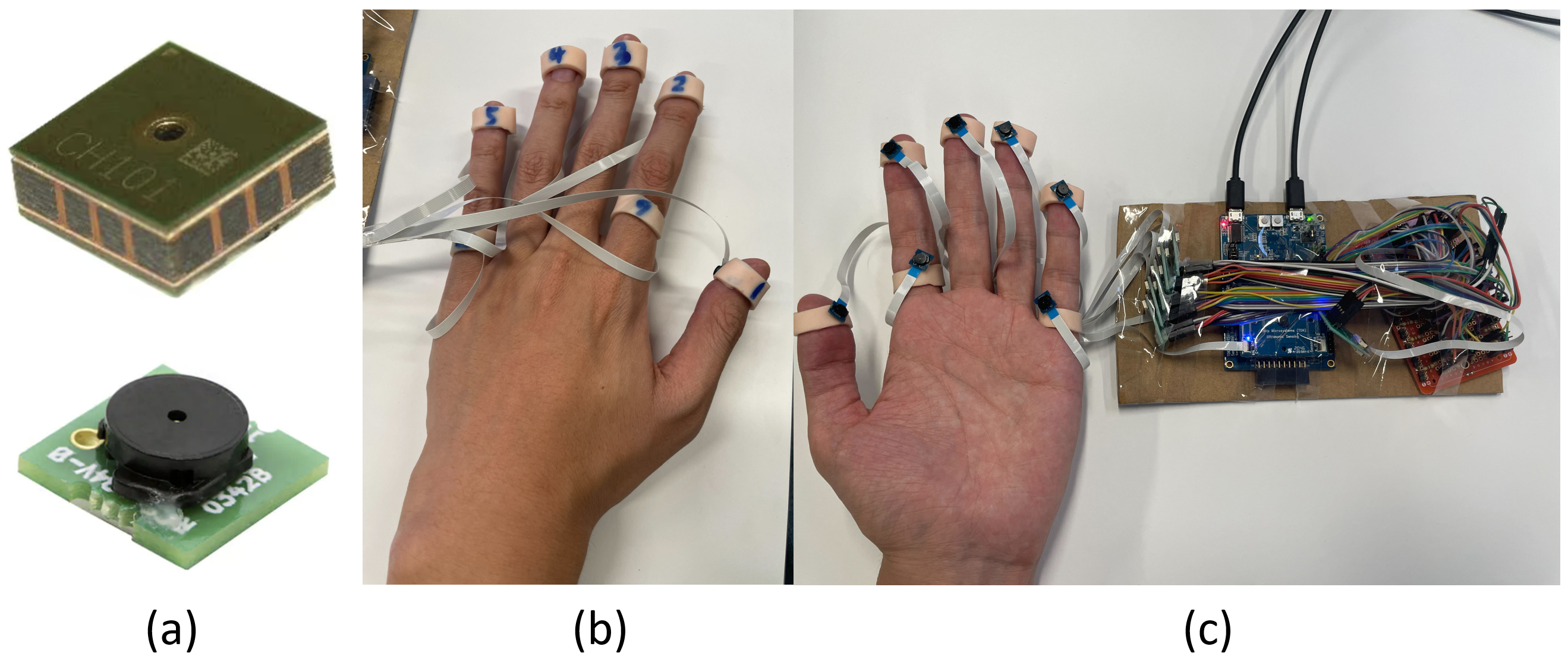}
  \caption{Visualization of sensors and how they are attached to the human hand in our system. From left to right: (a) Single CH101 sensor. (b) The back side of the hand with the sensors attached. (c) The front side of the hand and the embedded system circuit.}
  \label{fig:sensorattach}
\end{figure}

In our prototype, we attach 7 sensors to the hand as shown in Fig.~\ref{fig:sensorattach}. Subfigure (a) shows a single sensor, (b) demonstrates how they are attached, and (c) shows how the circuit and the sensors are connected.

\subsection{Sensor Data Acquisition System}
We choose 7 CH-101 sensors that communicate with the SmartSonic development board from TDK using the I2C protocol. During measurement, we cycle through sensors 1 to 7 to select one sensor at a time as the transmitter, while the remaining sensors serve as receivers. This approach enables us to obtain six distance values simultaneously and create a complete distance matrix within a single cycle. Then the development board relays the raw sensor matrix to the laptop using the serial protocol. 

\begin{figure}
\centering
  \includegraphics[width=0.9\columnwidth]{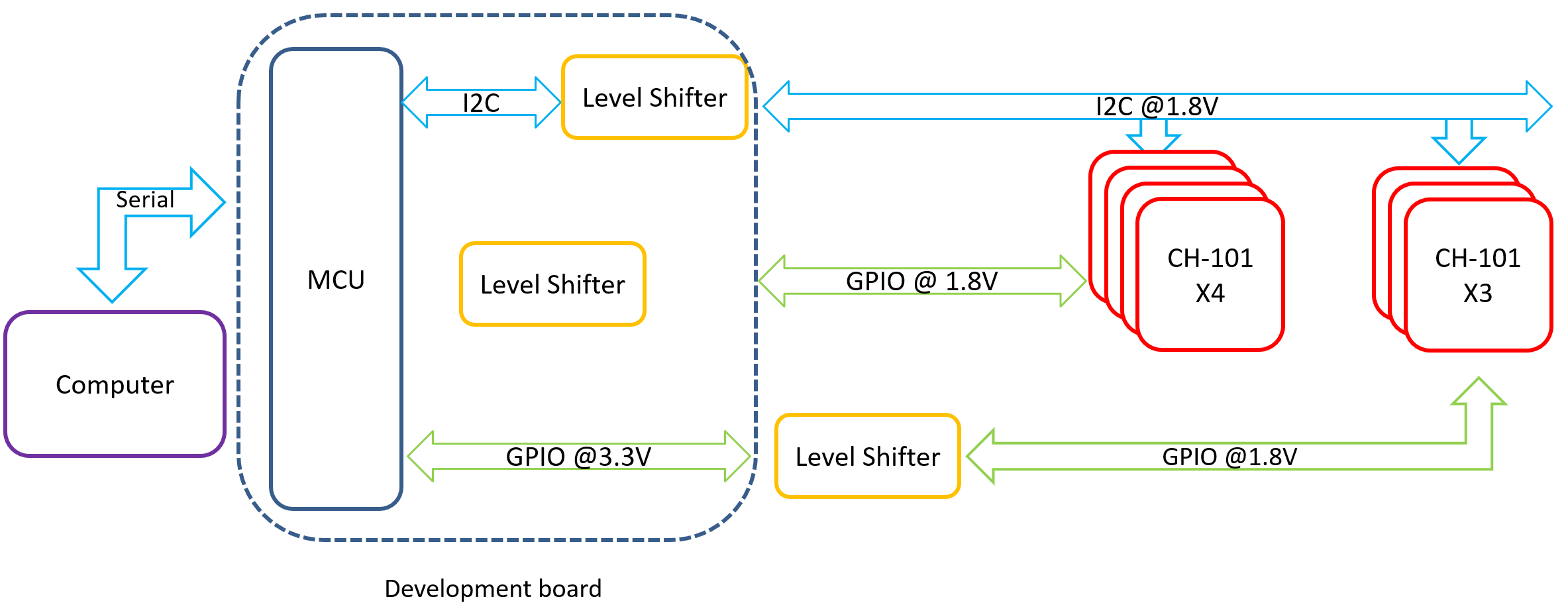}
  \caption{\finalrev{System-level diagram visualization for data acquisition with the SmartSonic development kit. The development kit provides an I2C bus to all 7 sensors at 1.8V. It also provides enough I/O pins that are internally level-shifted for the 4 ultrasound sensors. However, external level-shifters are used for the I/O pins of the remaining 3 sensors. The board uses a serial link to send sensor readings to the computer.}}
  \label{fig:devkit}
\end{figure}

\begin{figure*}
  \centering
  \includegraphics[width=1.70\columnwidth]{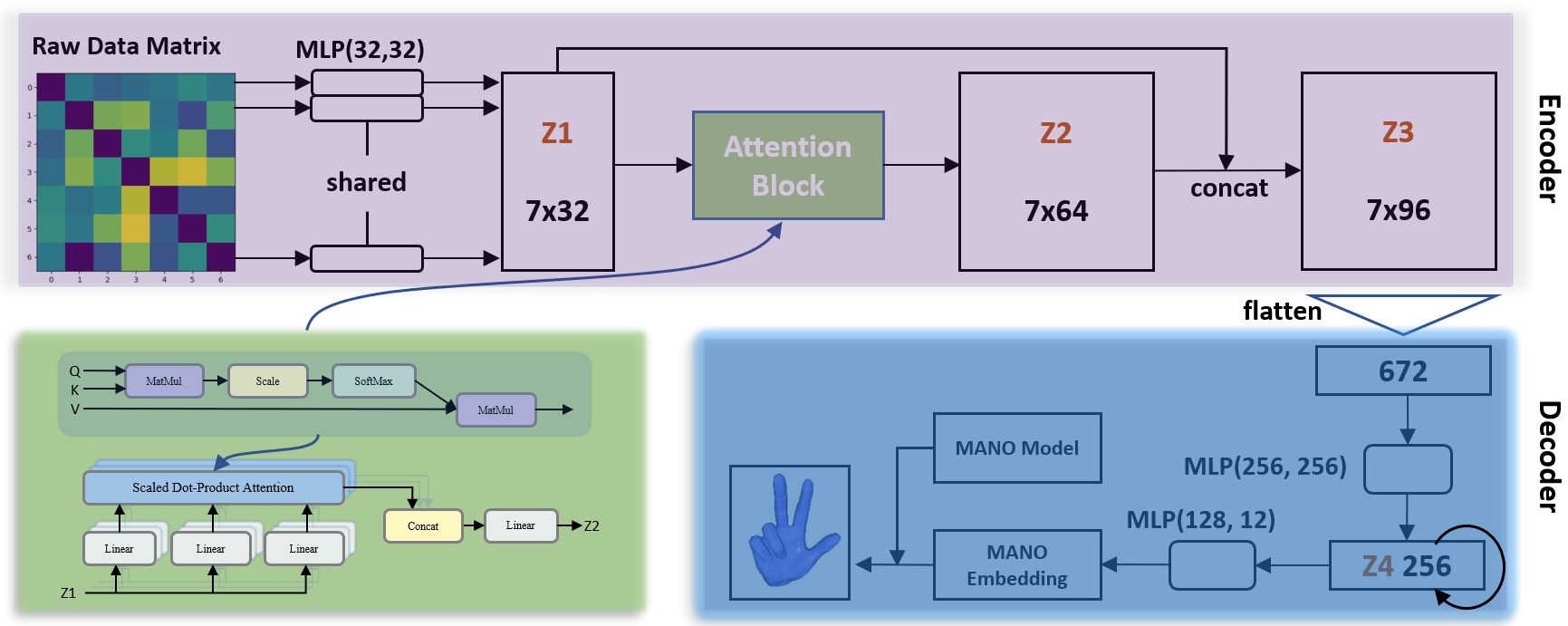}
  \caption{Pose Prediction Model Framework. Our model consists of an encoder and a decoder. The encoder module takes the raw data matrix as input, then feeds it into an MLP, followed by the attention block, whose output is concatenated with the MLP feature. For the decoder, we first flatten the feature and feed it into an LSTM to incorporate information from past frames and the MANO model to incorporate a hand pose prior. }
  \label{fig:figure2}
\end{figure*}

As shown in Fig.~\ref{fig:devkit}, the development board only provides enough level-shifted I/O ports for 4 sensors. Therefore, additional off-board level-shifting circuits are used to translate between 3.3V and 1.8V logic. For unidirectional buses, resistor dividers provide adequate performance due to the low acquisition rate in this system. A discrete-part translator from SparkFun (BOB-12009) is used to drive bidirectional pins. However, this system setup does not scale well due to development board limitations. We describe a scalable system that can support more CH-101s with a commercially available MCU in Section~\ref{sec:moresensors}.

\subsection{Encoder-Decoder Pose Prediction Model}
The low-level embedded system collects the raw pairwise distance matrix from the seven MEMS-based ultrasonic sensors, which is represented as a $7 \times 7$ matrix. This matrix is then fed into our pose prediction model, which predicts the hand pose represented by 23 joint positions. 
Our model, shown in Fig.~\ref{fig:figure2}, consists of an encoder with an attention mechanism and a decoder that incorporates information from previous frames as well as the MANO hand prior~\cite{romero2022embodied}. The encoder maps the $7 \times 7$ distance matrix into a $7 \times 96$ feature space, and the decoder uses this feature to predict the joint positions. Here we describe them in detail.


\paragraph{Encoder Module:} For each sensor, we pass its vector of distances to other sensors into an MLP ($7\rightarrow32\rightarrow32$) model, then stack the results to get a $7 \times 32$ feature embedding ($Z_1$). 

We then use a self-attention module to both provide robustness against missing measurements and extract the graph information among these sensors\,---\,intuitively, how these sensor distances formulate the hand pose pattern. To be specific, we use the classical multi-head attention model, built upon scaled-dot-product attention blocks:
\begin{equation}
\text{Attn}(Q,K,V)=\text{softmax}\left(\frac{QK^T}{\sqrt{D_k}}\right)V=AV,
\end{equation}
where $Q$, $K$ and $V$ are the query, key, and value, and $D_k$ is the dimension of the key.
In our model, we use different linear transformations of $Z_1$ for the query, key, and value in each attention head:
\begin{equation}
\text{head}_i = \text{Attn}\bigl(Z_1{W}_i^{{Q}}, Z_1{W}_i^K, Z_1{W}_i^{{V}}\bigr),
\end{equation}
where ${W}_i^{{Q}}$, ${W}_i^{{K}}$ and ${W}_i^{{V}}$ are learnable $32 \times 64$ matrices.
Finally, we concatenate these heads and multiply them with a final linear matrix $W^O$ to get the $7 \times 64$ feature embedding $Z_2$:
\begin{equation}
Z_2=\text{MH-Attn}(Q,K,V)=\text{Concat}(\text{head}_1,\text{head}_2,\dots)\,{W}^{{O}}.
\end{equation}
We concatenate $Z_1$ and $Z_2$ with a skip-connection to get the final encoded feature $Z_3$ with shape $7 \times 96$:
\begin{equation}
Z_3=\text{Concat}(Z_1,Z_2).
\end{equation}

\paragraph{Decoder Module:} We first flatten $Z_3$ into a one-dimensional vector, which is then followed by another MLP ($672\rightarrow256\rightarrow256$) to convert it into $Z_4$, with size 256:
\begin{equation}
Z_4=\text{MLP}\bigl(\text{Flatten}(Z_3)\bigr).
\end{equation}

To aggregate information from previous time steps, we use an LSTM model with the hidden dimension the same as the input dimension 256. The LSTM cell takes as input a sub-sequence of feature vectors $Z_4^1, Z_4^2, \ldots, Z_4^T$, where $T$ is the length of the sub-sequence. 
For each sub-sequence, the LSTM processes each feature vector $Z_4^i$ in order and updates its internal state. After the last feature vector in the sub-sequence is processed, the final hidden state of the LSTM is used as a summary and represents the aggregated information from the previous five timesteps:
\begin{equation}
F=\text{LSTM}\bigl(Z_4^1, Z_4^2, \ldots, Z_4^T\bigr), T=5.
\end{equation}



The MANO (Model for Articulated Hands) model~\cite{romero2022embodied} is a parametric 3D hand model that represents the human hand as a set of articulated bones, joints, and skin. It can be used to generate 3D hand poses from a set of input parameters. During the MANO hand training phase, a large amount of hand pose data is collected and subjected to PCA analysis, resulting in a set of principal component vectors. These principal components represent the patterns of variation in hand poses (joint angles). By adjusting the weights assigned to these principal components, different joint angles can be generated.


The weight parameter dimension of this MANO model is a hyperparameter, and here we set it as 12. We feed the feature information from the temporal LSTM module described above into an MLP ($256\rightarrow128\rightarrow12$) and pass its output to the MANO hand model:
\begin{equation}
J=\{J^1,J^2, \ldots, J^{n}\}=\text{MANO}\bigl(\text{MLP}(F)\bigr),
\end{equation}
where the $J^i$ are the parameters of the joints and $n$ is the number of degrees of freedom: 23 for the human hand, though Sec.~\ref{sec:mechhand} presents an experiment with a mechanical hand with $n=5$ DOF.

Our entire model\,---\,encoder and decoder\,---\,is trained end-to-end with L2 loss.  Training takes 1-2 hours on a single Nvidia 3080Ti.

\subsection{Sim-to-real Training Pipeline}
To achieve the best possible performance in hand pose estimation, we adopt a sim-to-real transfer training pipeline. This pipeline involves several steps. First, we consider the motion of a simulated hand, and compute sensor positions relative to that hand in the same configuration as we use in our real-world physical setup. This ensures that the simulated data captures the same physical interactions between the hand and the sensors as in the real world.

Secondly, we position the simulated hand using sequences of poses from the InterHand2.6m dataset~\cite{moon2020interhand2}, and compute the distances between the simulated sensors.  We augment the distance measurements by adding noise and randomly mask measurements to simulate missing data.  This allows us to generate a large amount of labeled training data in a controlled and reproducible way that nevertheless resists overfitting and accommodates a wide range of hand sizes and measurement imperfections.

Next, we train a sequential pose prediction model using the simulated dataset. This model takes the sequence of hand poses as input and predicts the next pose in the sequence. By training on the simulated data, the model learns to generalize well to variations in hand shape and movement.

Finally, we adapt the model to the real-world domain by fine-tuning it on a real dataset (described below).
This sim-to-real transfer training pipeline has been shown to be effective in improving the performance of hand pose estimation models, especially in scenarios where large amounts of labeled real data are not available.

\begin{figure}
\centering
  \includegraphics[width=0.95\columnwidth]{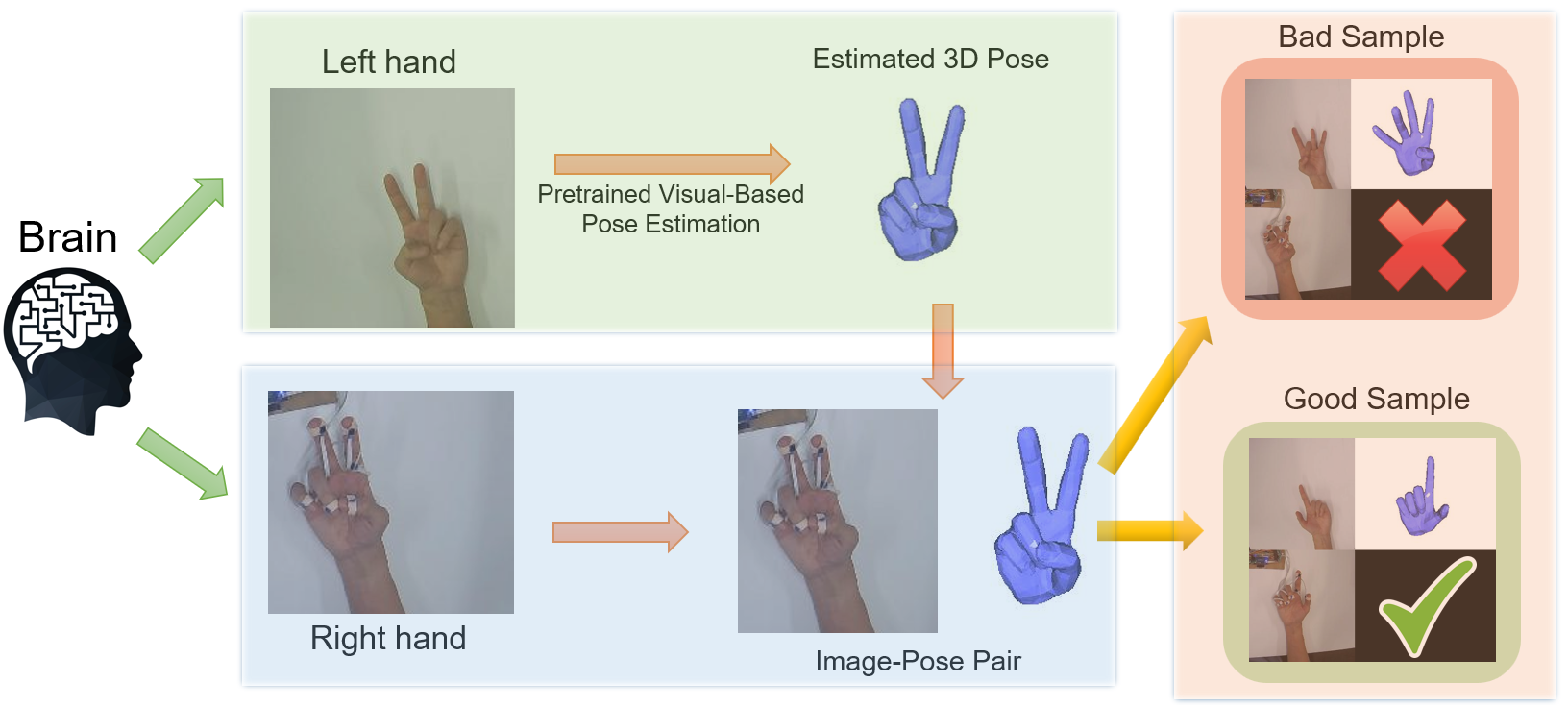}
  \caption{Dataset collection system visualization. In this figure, we visualize how to collect the raw image and obtain its (pseudo-)ground truth pose via left-right-hand synchronization, a pretrained hand pose estimation model, and the human filtering process. The visualization of hand poses is derived from ~\cite{romero2022embodied}.}
  \label{fig:collection}
\end{figure}

\subsection{Dataset and Pseudo-Ground Truth Collection}

\paragraph{System Setup:} We obtain a dataset, used for both fine-tuning our model and qualitative evaluation, consisting of raw sensor distance data synchronized with a pseudo-ground truth obtained using a vision system. To account for processing delays that can occur when collecting data in a single process, we adopt a multi-process collection approach, in which one process handles the raw data and another computes the ground truth from a camera video stream. These two processes each use their own data buffers.


\paragraph{Pseudo-Ground Truth Extraction:} It is not easy to manually label each hand pose from scratch, nor to perform visual tracking on a hand with our sensors attached to it.  Nevertheless, we find that our training procedure benefits from a small amount of real-world fine-tuning, even with imperfect ``ground truth.''  We therefore choose a procedure that sacrifices accuracy in the ground truth in return for ease of acquisition.  Specifically, we mount our sensors to a person's right hand and ask them to perform a set of hand poses while mirroring what they do with their left hand, using the latter (with a mirror reflection) for our vision-based system ~\cite{li2022interacting} (see Fig \ref{fig:collection}).  This introduces the possibility of several kinds of error, most notably that the vision system fails or that the left- and right-hand poses are not synchronized.  To combat the former, we manually screen the output of the vision system and remove any outputs that do not appear to match the video frames.  For the latter, we remove frames in which any finger's position differs by more than 4mm between the pre-trained estimated pose and the vision pose.  In all, we end up filtering out approximately 15\% of frames, which still leaves us with a dataset that is useful for fine-tuning.

\paragraph{Hand Position and Orientation Normalization:} Since our system does not predict global hand position and orientation, we remove this information from the dataset by normalizing each hand pose.  We first shift the whole hand such that the wrist point lies at the origin, then rotate the hand such that the root of the middle finger is located on the Z-axis and the root of the index finger is located on the X-Z plane.

\section{Experimental Verification and Applications}

\subsection{Dataset Statistics}\label{sec:dataset}
We use the InterHand2.6m dataset~\cite{moon2020interhand2} for pre-training. This is a large-scale hand pose estimation dataset containing over 2.6 million hand images with corresponding 3D hand joint annotations. 
Although there are 2.6m images, they represent multiple views for each of around 46k hand poses. We use all of these 46k hand poses as the simulation dataset.  Some pose samples are visualized in Fig.~\ref{fig:raw}a. 

Our real-world dataset contains around 5000 frames, each containing a raw distance matrix and the hand pose represented as 23 joint positions. For the distance matrix, when one sensor misses the signal sent from another sensor, we mark the value as -1. This missing data accounts for less than 1\% of the whole dataset. Fig.~\ref{fig:raw}b shows sample frames from our real-world dataset, while Fig.~\ref{fig:raw}c visualizes the corresponding distance measurements.

\begin{figure*}
  \centering
  \includegraphics[width=0.95\textwidth]{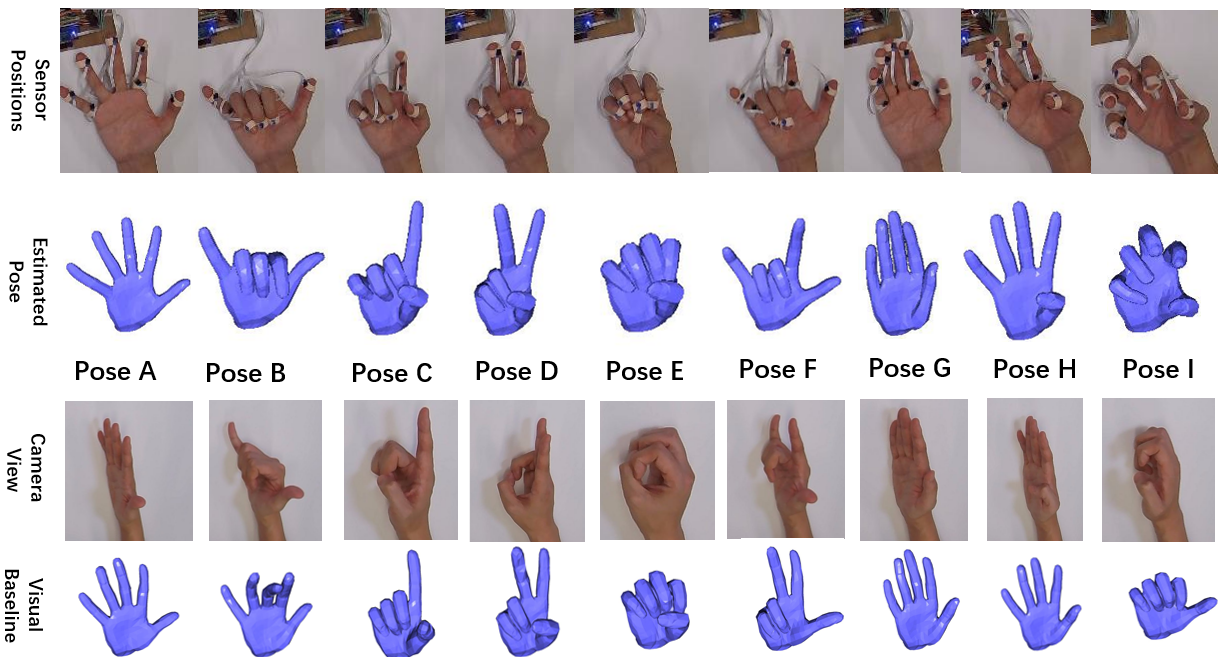}
  \caption{Qualitative Performance Evaluation. From top to bottom, we show a hand with our sensors attached, the estimated hand pose, the mirrored camera view image for the same pose, and the vision baseline results. Our method qualitatively outperforms the vision baseline, especially in the presence of occlusion. }~\label{fig:quali}
\end{figure*}

\subsection{Raw Sensor Accuracy Analysis}
\begin{figure}[h!]
\centering
  \includegraphics[width=0.95\columnwidth]{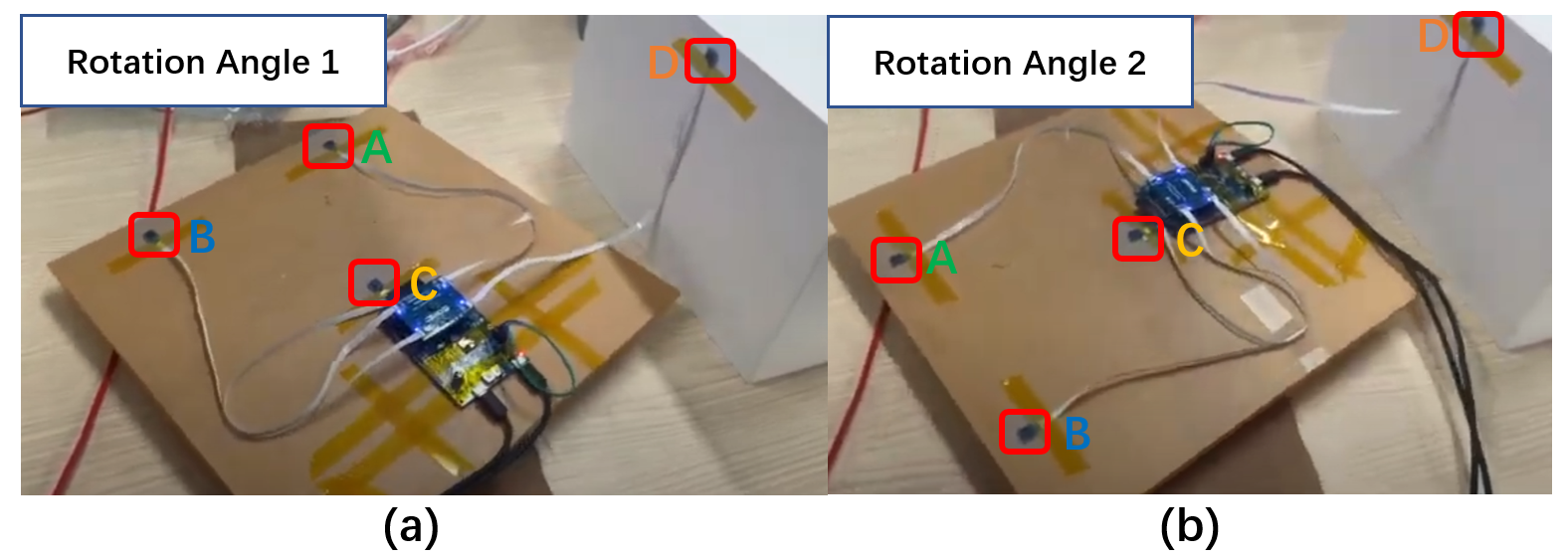}
  \caption{Evaluation of localization accuracy. Sensors A, B, and C are attached to the rotating platform, while sensor D is fixed on the nearby box. (a) and (b) show the setup at two different angles of rotation. }
  \label{fig:3p}
  \vspace{1em}
%
\centering
  \includegraphics[width=1.0\columnwidth]{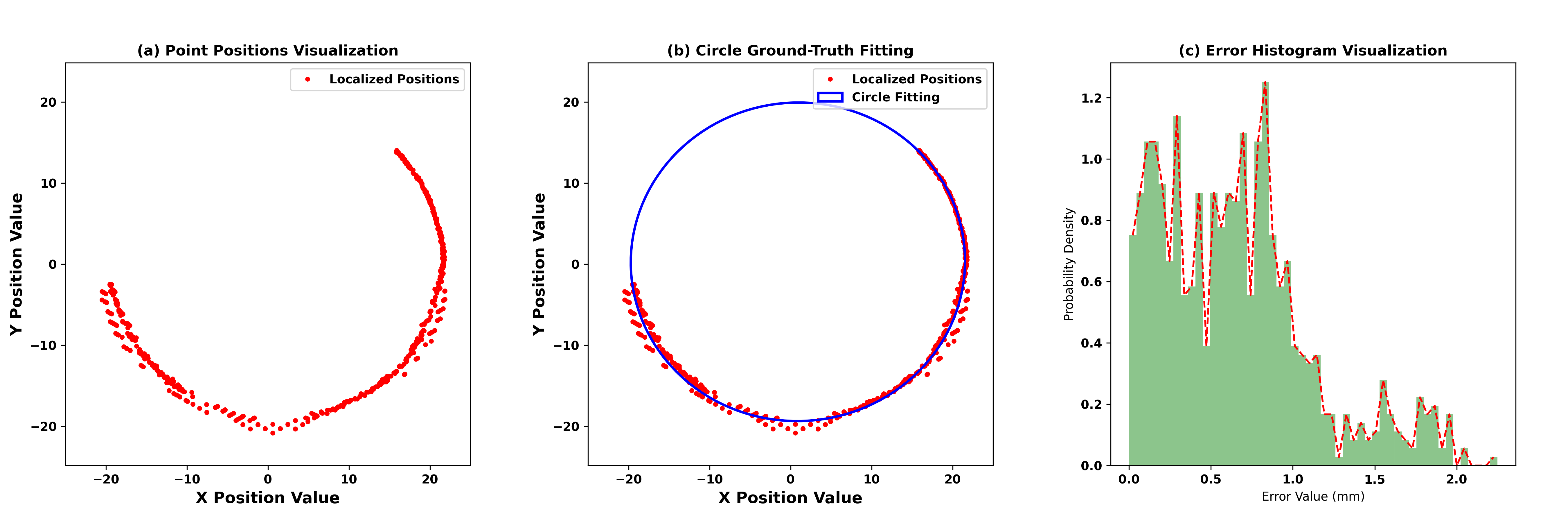}
  \caption{The sensor D is localized relative to a coordinate system defined by A, B, and C.  (a) Scatterplot of the coordinates of D as the platform is rotated.  (b) Best-fit circle, showing the qualitative accuracy of localization.  (c) Histogram of distances between points and the best-fit circle.}
  \label{fig:3p_traj}
\end{figure}

Here we describe an experiment to evaluate the accuracy of the raw sensor data. As shown in Fig. \ref{fig:3p}, we attach three sensors (named A, B, and C) at the vertices of an equilateral triangle on a rotating platform. There is also another sensor D attached to the nearby box, and the box is always fixed. We then rotate the platform and collect the sensor distances between D and A, D and B, as well as D and C. We compute the analytical position for sensor D based on its distances to other sensors, in a coordinate system centered at C and with the direction from B to A as the x-axis and the vertical direction as the z-axis.  We thus expect the computed coordinates of D to lie in a circle, as the platform is rotated.

The point positions projected to the X-Y plane are shown in Fig.~\ref{fig:3p_traj}a, and we can fit a circle to these points as shown in Fig.~\ref{fig:3p_traj}b. It is qualitatively clear that all the points lie close to the circle. Quantitatively, the average localization error is 0.65mm, and most errors are below 1.2mm, as shown in the histogram in Fig.~\ref{fig:3p_traj}c. 

\subsection{Qualitative Evaluation on Human Hand Poses}
We train our model on real human hands using the sim-to-real pipeline described above, and use it to visualize the qualitative performance of our system.  Fig.~\ref{fig:quali} shows a series of poses, with an image of the hand wearing the sensors at top and our predicted hand pose (rendered in the Open3D engine, with the MANO hand representation~\cite{romero2022embodied}) below.  We compare our results to a vision baseline ~\cite{li2022interacting}, demonstrating the frequent inaccuracy in the latter, especially for poses that incorporate significant occlusion with respect to the camera.

\paragraph{Size-agnostic Hand Pose Estimation:}
Our pose estimation model is designed to be agnostic to hand size, given that it is trained on a synthetic dataset containing hands of multiple shapes. Figure \ref{fig:size} evaluates hand poses estimated for multiple individuals with hands of different sizes and shapes. The top row shows results for the same person on which the model was fine-tuned, while the other rows demonstrate good adaptability to individuals with smaller hands. \finalrev{The hands from top to bottom have lengths of 21.4 cm, 19.2 cm, and 15.5 cm, and widths of 8.1 cm, 7.7c m, and 7.0 cm. Given the large variation in hand sizes, the system generalizes well though not perfectly.}

\finalrev{
\paragraph{Pseudo Ground-truth Quantitative Comparisons:}
The average distance between our predicted joint positions and the pseudo ground-truth joint positions, as measured on the test set, is 1.09 cm. Although we cannot determine how much of the error comes from the pseudo ground truth error vs.\ our proposed system framework/model, it demonstrates that the error level of our system is relatively low. 
\paragraph{Nearest Neighbor Baseline:}
We also provide the results for nearest neighbor baseline: during the inference stage, we measure the similarity between the real-time distance matrix from the embedded system and the dataset distance matrices, and select the hand pose with the maximum similarity to represent the model prediction. Here the similarity is defined as the cosine distance between the two 49-d vectors(flattened from 7x7 matrix). It turns out that the quantitative result for this baseline is 3.64cm, which is much higher than ours and demonstrates the effectiveness of our model.
}

\begin{figure}
\centering
  \includegraphics[width=0.8\columnwidth]{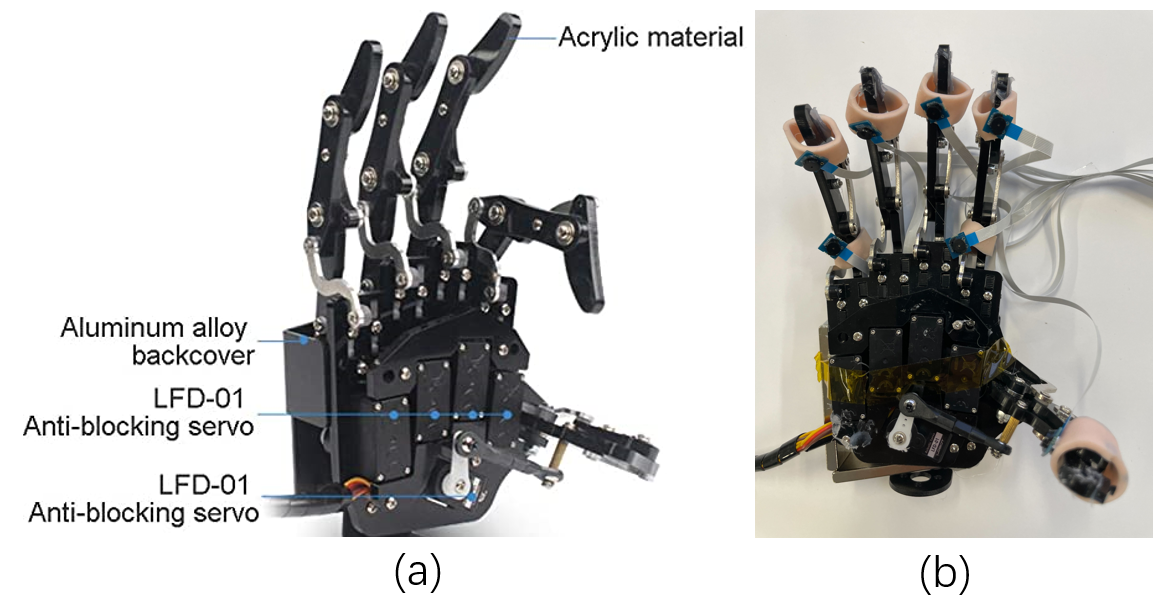}
  \vspace{-0.5em}
  \caption{(a) Mechanical hand with five degrees of freedom, used for quantitative evaluation. (b) Hand with sensors attached, in the same configuration as that for the human hand.}
  \label{fig:mechhand}
  \vspace{1em}
%
\centering
  \includegraphics[width=\columnwidth]{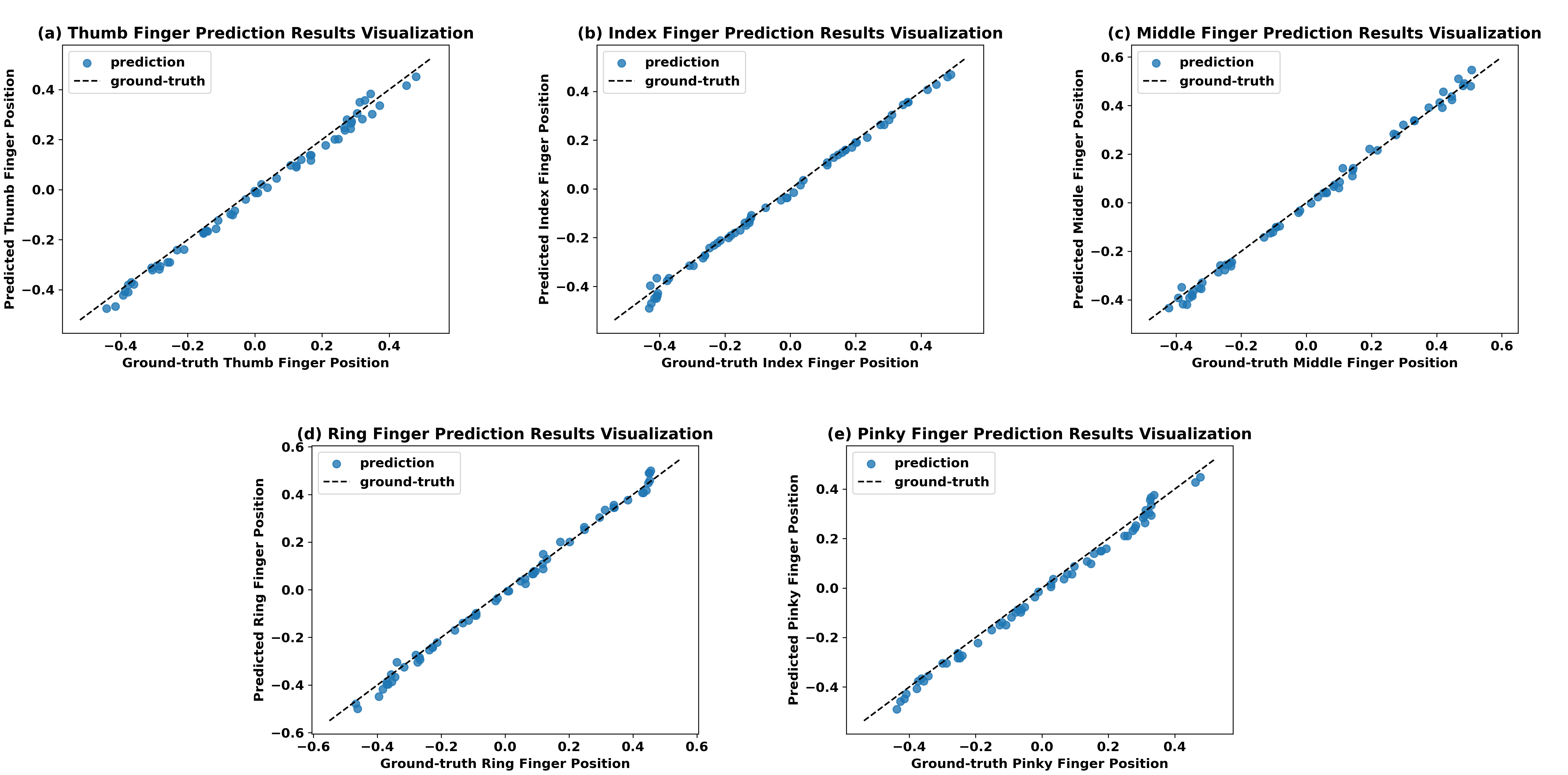}
  \caption{Results of accuracy experiment using mechanical hand.  The graphs show results for the thumb, index, middle, ring, and pinky fingers, respectively. The horizontal axis is the ground-truth servo command, while the vertical axis represents the inferred servo command.}
  \label{fig:mechresults}
  \vspace{-0.5em}
\end{figure}

\subsection{Performance Evaluation Using a Mechanical Hand} \label{sec:mechhand}
To quantitatively evaluate our system, we conducted experiments using a mechanical hand (Fig.~\ref{fig:mechhand}) with five degrees of freedom. Each finger is controlled by a separate servo motor, and we can therefore define a 5-dimensional space of poses parameterized by the 5 servo commands.

To accommodate the difference between the 23-dimensional human hand model and the 5-DOF mechanical hand, we remove the MANO stage from the original model and replace it with an MLP ($256 \rightarrow 128 \rightarrow 5$) model, which directly regresses the five-finger command signals.

Our dataset consists of 30,000 frames, each including a raw data distance matrix and the corresponding hand servo commands representing the hand pose.
Fig.~\ref{fig:mechresults} shows results from a testing dataset, comparing actual and predicted poses.  In each graph, the horizontal axis denotes the ground truth pose value (normalized to the range $[-0.5, 0.5]$), while the vertical axis represents our prediction result. The mean error was found to be 0.0163, demonstrating that our model can achieve excellent performance.





\section{Discussion}

\subsection{Ablation Study of Our Model}
Here we provide an ablation study for the model we designed for the hand pose estimation to illustrate the effectiveness of each module. As shown in Table~\ref{tab:ablation}, these three ablation study experiments represent removing the sequential module, attention module and skip connection.  We find that all three of these modules contribute to the final performance of our system.

\finalrev{We also provide an ablation study for the fine-tuning sim2real stage. Recalling the pseudo ground-truth quantitative metric, fine-tuning reduced the error from 1.45 cm to 1.09 cm, thus demonstrating nontrivial improvement. This effectiveness comes from differences between the simulator and the real-world dataset: 1. Slightly different sensor positions between the simulator and the real world. 2. Different coverage of hand poses between simulation and the real world. 3. Lack of physical effects such as ultrasound reflection or transmission through occluders in the simulation. It would be possible but not easy to add these to our simulator, since they would require significant calibration against the real world.
}

\begin{table}[h]
\caption{Ablation study of components in our pose prediction model.  The full model with sequential processing, attention mechanism, and skip connection provides the highest accuracy.}
\begin{tabularx}{\columnwidth}{XXXXX}
\toprule
 & w/o seq. & w/o atten. & w/o skip & full \\
 \midrule
L2 loss & 0.0196 & 0.0215 & 0.0207 & \textbf{0.0163} \\
\bottomrule
\end{tabularx}\label{tab:ablation}
\end{table}

\balance

\subsection{Alternative Ultrasonic Sensor Selection}\label{sec:otherultra}
\begin{figure}[h!]
\centering
  \includegraphics[width=0.95\columnwidth]{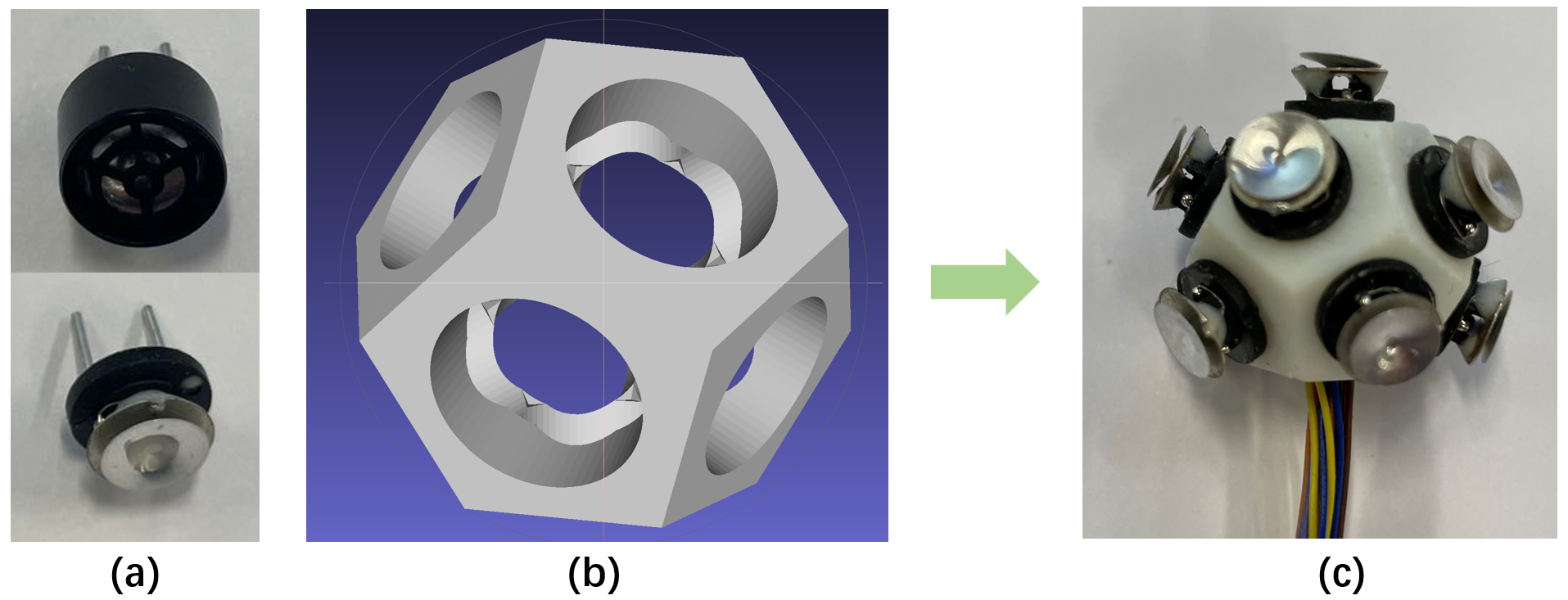}
  \caption{Dodecahedral design for omnidirectional sensing using low-cost, narrow-beam piezoelectric ultrasonic transducers. (a) Individual sensor with and without its casing. (b) 3D CAD design for the support frame. (c) Assembled sensor array. }
  \label{fig:sphere}
\end{figure}

\begin{figure*}[p]
\centering
  \includegraphics[width=0.9\textwidth]{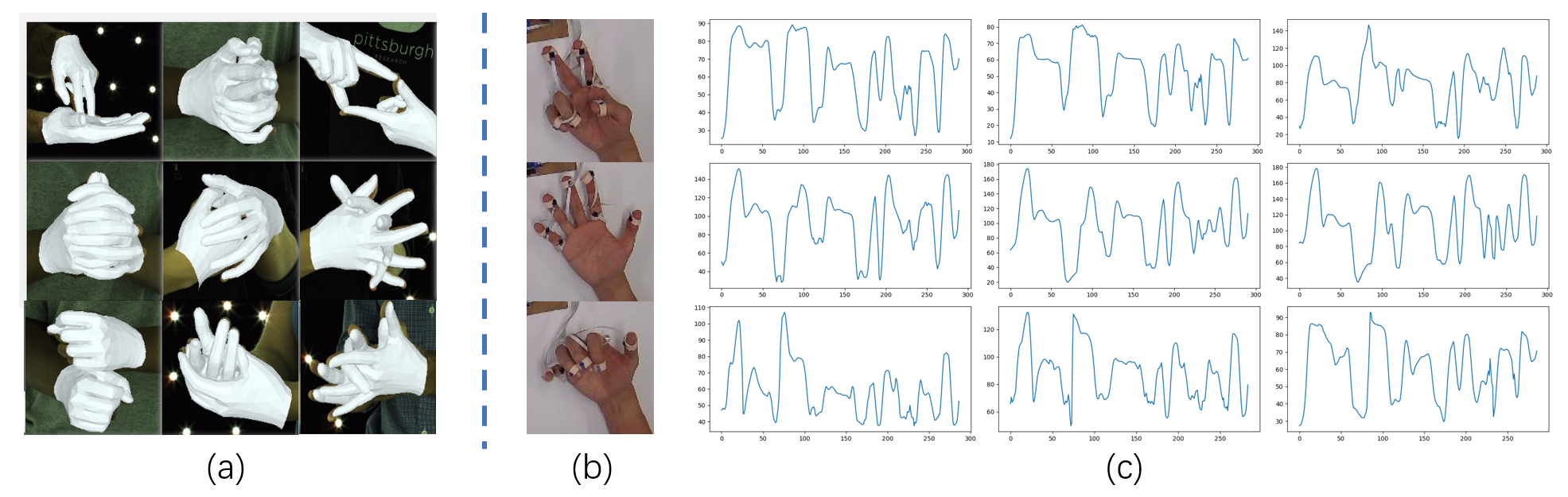}
  \vspace{-0.5em}
  \caption{Dataset Visualization. (a) The InterHand 2.6M dataset used for pre-training. (b) Hand poses for fine-tuning.  (c) Raw sensor data displaying variation as the hand moves to different poses. There are three subfigures in each line, representing the data feature dimensions 1-7, 8-14, and 15-21.}
  \label{fig:raw}
  \vspace{1.5em}
%
  \centering
  \includegraphics[width=0.93\textwidth]{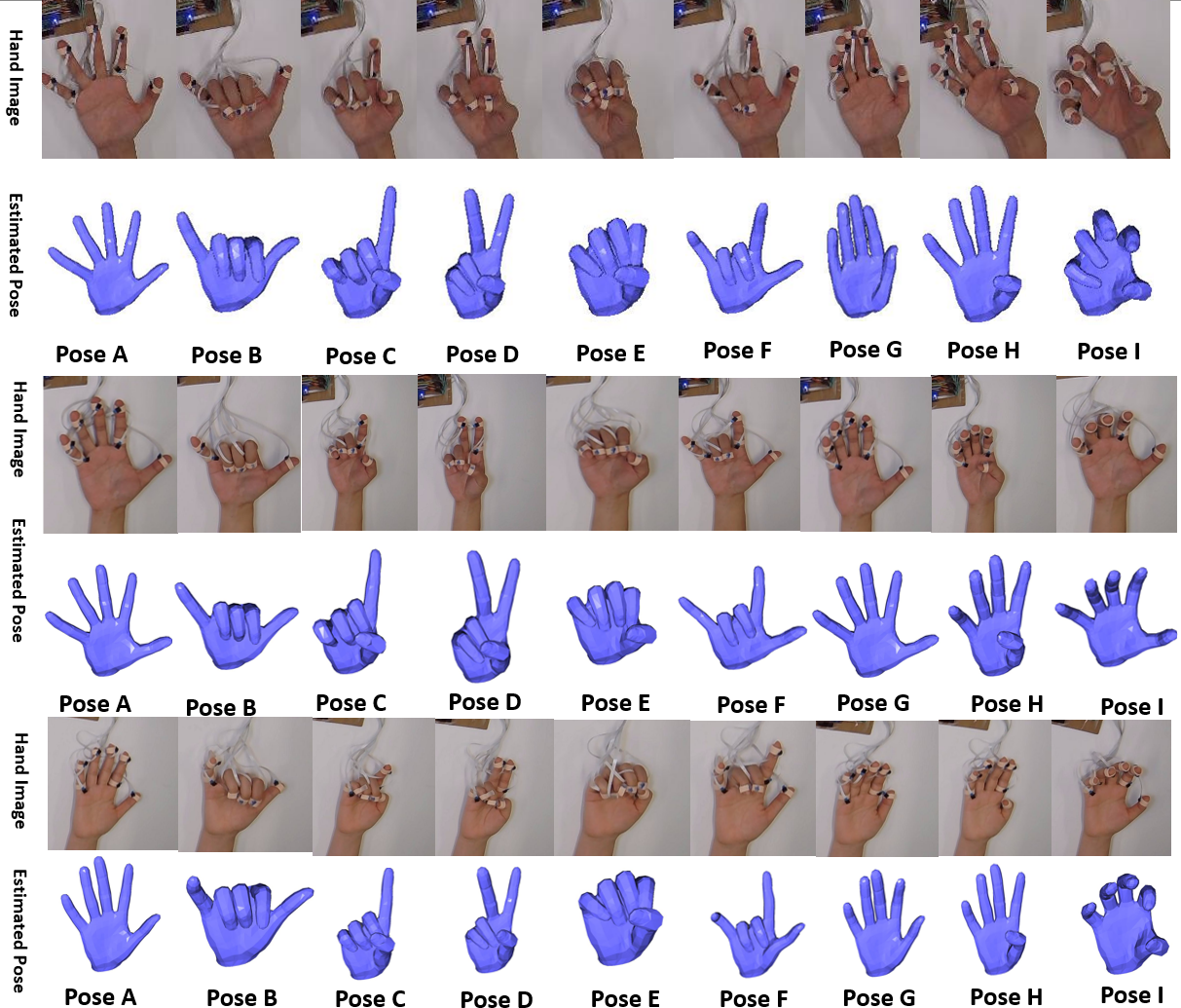}
  \caption{Evaluation of the sensitivity of results to hand size and shape. Here we show that the model fine-tuned on one dataset can be adapted to hands with different sizes and shapes. The top result is for the same hand size (``large'') as the data collected for fine-tuning, while the middle (``medium'') and lower (``small'') results show qualitatively similar performance for individuals with different hand sizes.}
  \label{fig:size}
  \vspace{-2em}
\end{figure*}

We also experimented with another, more traditional type of ultrasonic sensor that is based on the piezoelectric effect. As shown in Fig~\ref{fig:sphere}a, the upper image is the individual sensor and the lower one is the sensor without its outer casing. We have removed the casing to enable transmission and reception across a wider beam width. However, the improvement in beam angle was not sufficient, and we found that we could not receive signals emitted from beyond approximately 60 degrees. Consequently, we designed a dodecahedron-shaped support frame as shown in Fig~\ref{fig:sphere}b and 3D-printed this support frame, which allows the placement of 12 sensors. This dodecahedral sensor array is omnidirectional for both transmitting and receiving ultrasound waves. Since the array can be driven directly from the MCU ports without an intermediate translator or driver, a refresh rate of up to 500Hz was achieved.

However, despite the potential benefits in refresh speed, omnidirectionality, and cost, we ended up not using this sensor configuration based on two drawbacks.  First, the assembled array is too large, with a radius above 15mm.  While this may not be a problem for other tracking applications, such as body pose capture, the size rendered it unfit for attachment to fingers.  Secondly, the distance resolution is relatively low and the noise level is higher than the MEMS-ultrasonic sensors due to low ultrasound frequency.

\subsection{Different Sensor Configurations Design}
We also experimented with several different sensor configurations in our system. The number of sensors can be varied from 5 to 8 or more. Here we analyze the trade-offs between different sensor configurations. The minimum number of sensors is 5, and we attach them at the fingertips. The number of pairwise distances available from these sensors is only 10, which is lower than the number of degrees of freedom of a human hand. Moreover, this minimal number of sensors provides very little robustness to missing measurements due to occlusion.  If we add one more sensor, the best place to put it is on the wrist. However, this sensor attachment method is not stable due to the movement of the wrist. For seven sensors, we place the two additional sensors at the root of the index and little finger for optimum performance.  The performance gain plateaus with additional sensors beyond seven regardless of the sensor placement.

\finalrev{Quantitatively, we obtain simulated mean errors of 1.24 cm, 1.07 cm, 0.85 cm, and 0.82 cm for 5-8 sensors, respectively. The results match with the real experiment results. Based on these analyses and experiments, we motivate our conclusion that a 7-sensor setup offers the best tradeoff between accuracy and being easy to wear.}

\subsection{Different MCU and Embedded System Design}\label{sec:moresensors}
 In this section, we propose a general framework for expanding the number of supported sensors. Though developed mainly for hand motion capture, this system has the potential to scale to a variety of motion capture settings, such as wrist, arm, torso, or whole-body movement. Therefore, sensor scalability is an essential issue for the universality of the algorithm. The development kit used to demonstrate the hand-tracking system in this work only supports up to 8 sensor nodes. To achieve a wider range of motion capture applications, more sensors are needed to maintain the spatial and temporal resolution of the dataset. We present a solution for scalable deployment for higher numbers of sensor nodes in the supplemental material.  Dramatically increasing the number of sensors will also require solutions such as frequency multiplexing to avoid linear growth in the time of each measurement cycle.


\section{Conclusion}
We propose a novel hand motion capture glove based on MEMS-ultrasonic sensors.  Our work represents a non-trivial improvement in the field of hand-tracking, as it addresses the limitations of existing solutions and provides a practical and low-cost alternative for accurate and robust hand pose estimation. The proposed design and methodology can be applied to various applications such as virtual reality, human-computer interaction, and dexterous robot manipulation. The main limitations of our system include our real-world fine-tuning, which relies on mirrored hand motion, as well as the fact that our accuracy degrades when an object is held in the hand, since this blocks the propagation of ultrasound waves. However, this can be solved by attaching more sensors, with some on the front of the hand and others on the back side of the fingers\,---\,this is a possible direction for future work. Another avenue of future work includes extending the current framework to full human body pose estimation.

\clearpage
\bibliographystyle{ACM-Reference-Format}
\bibliography{sample}

\clearpage

\clearpage

\section*{Supplemental Material }
\vspace{0.5in}
\appendix

\section{Towards full-duplex operation for high refresh rate}
Real-time motion capture for precision movement reconstruction requires a high capture rate. With the advance of CMOS image sensors and high-speed photography, optical-based motion tracking has achieved a refresh rate in excess of 200 fps (frames per second) \cite{9220075, eventCap2020CVPR, 10.1145/3272127.3275062}. \finalrev{The ultrasound system demonstrated in this work has a refresh rate of around 10~Hz. The relatively low rate is due to delays in interrogating and waiting for replies from each sensor. To increase the refresh rate of the sensor front-end, FPGAs or dedicated controllers can be used to create multiple I2C buses for paralleled readback operation instead of sequential polling.} Even faster operations can be achieved with a concept similar to that of FDMA. By having sensors sensitive to a wide spectrum of frequencies instead of one particular frequency, Therefore, all sensors can interrogate different frequencies simultaneously, allowing full-duplex operation and increase refresh rate significantly. 

 \begin{figure}[h!]
\centering
  \includegraphics[width=0.85\columnwidth]{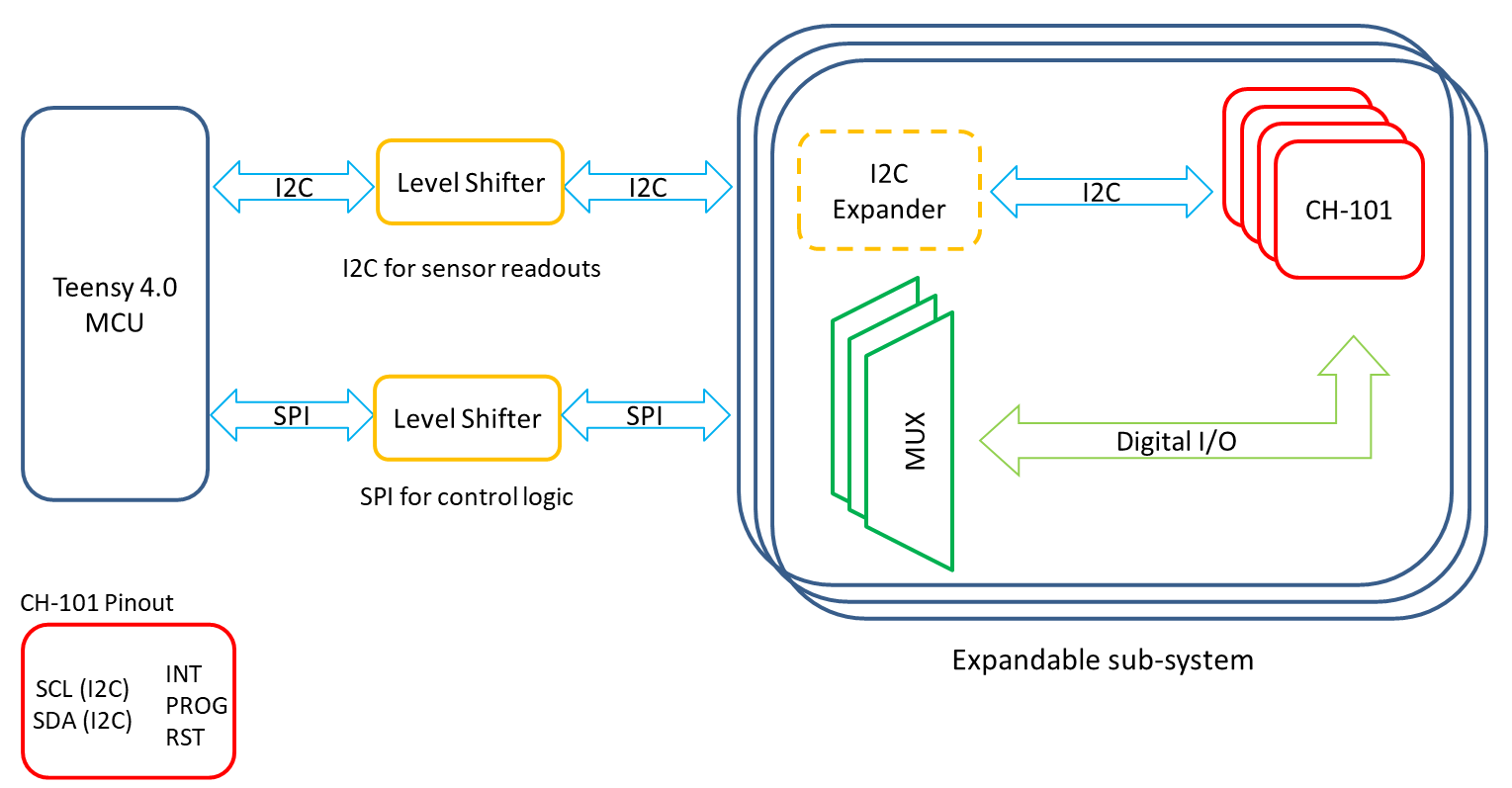}
  \caption{System-level diagram for commercial MCU adaptation}~\label{fig:teensy}
\end{figure}


\begin{figure*}[h!]
  \centering
  \includegraphics[width=1.80\columnwidth]{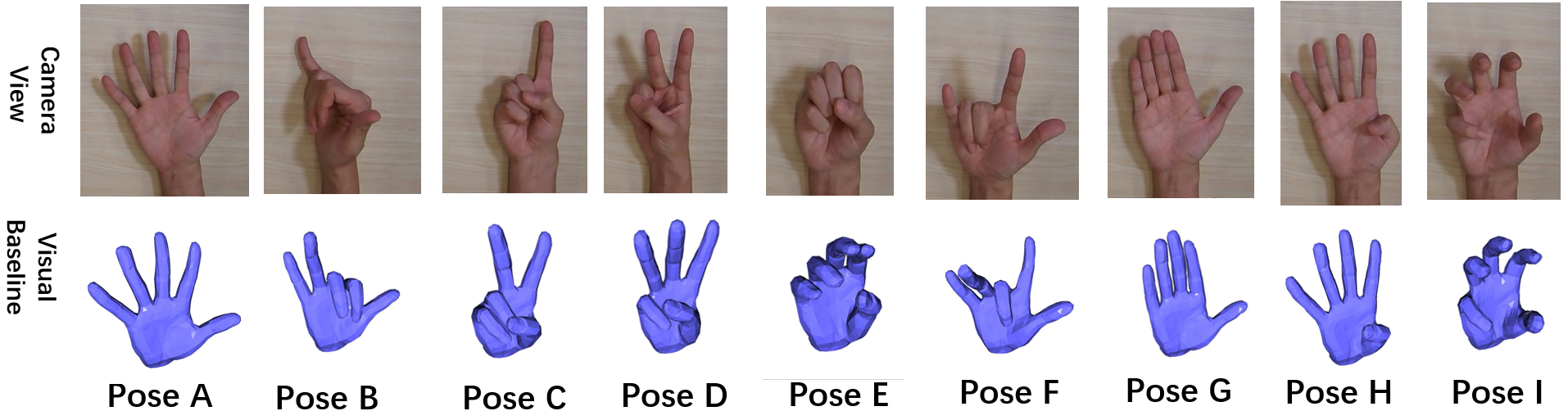}
  \caption{More visualization for visual-based baseline results. This visual-based algorithm is easily collapsed when the background color is close to the hand color. Meanwhile, the model is sensitive to the light conditions.}~\label{fig:morebaseline}
\end{figure*}




\section{Additional comparisons with the vision baseline}
We provide more analysis with the pure visual-based hand pose estimation model~\cite{li2022interacting}, as shown in Fig \ref{fig:morebaseline}.  It is apparent that the visual-based algorithm is sensitive to the background and the lighting conditions. When the background color is close to the human hand color, the model is collapsed.

\section{Towards scalable sensor acquisition system adaptation }
 
 Unlike the majority of commercially available sensors, CH-101 only supports 1.8V logic, while many commercial MCUs work at 3.3V or higher. Additionally, CH-101 uses a non-standard bi-directional drive mechanism that is not compatible with the popular push-pull or low-side open-drain counterparts. Special considerations are needed to address these problems.

 As shown in Fig. \ref{fig:teensy}, the proposed system contains multiple sub-units with each sub-unit supporting multiple CH-101 sensors. Each sub-unit consists of digital Muxes for logical I/O controls and an I2C expander for data readouts. Then each sub-unit's I2C and SPI buses are level shifted to 3.3V to match that of MCUs. Due to I2C buses' speed restriction (100khz typical or 400khz with high-speed mode), SPI-operated MUXs, which have a bus speed on the order of MHz, are chosen to handle control I/Os to avoid occupying the bandwidth. The optional I2C expander avoids the problem of multiple slaves having the same address and allows for another type of sensor to share the same I2C bus. The number of sensors each sub-unit supports depends on the MCU speed and I2C protocols. For higher-speed operations, an FPGA is preferred due to its reconfigurity.\par

\begin{figure*}[h!]
\centering
  \includegraphics[width=1.4\columnwidth]{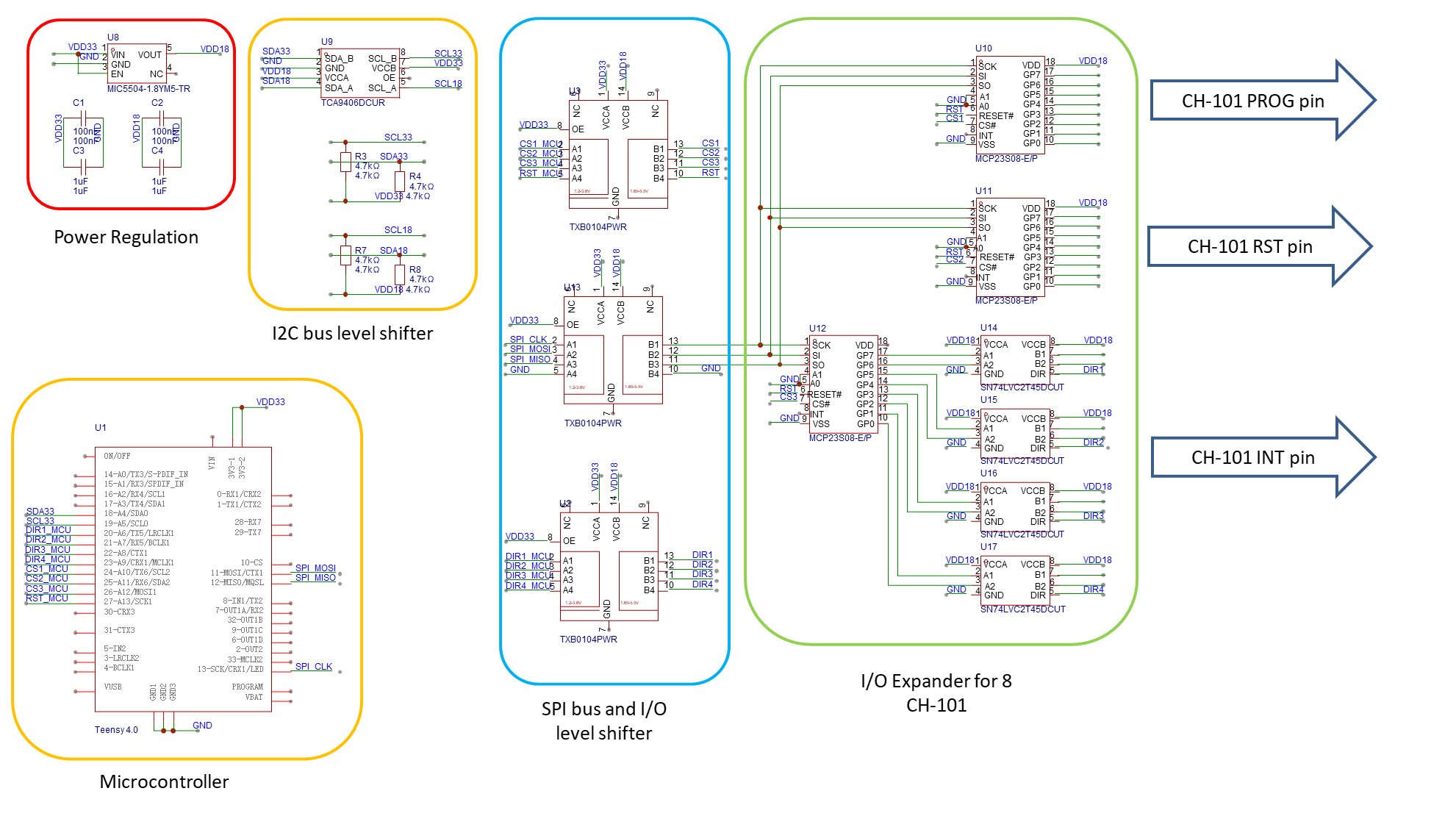}
  \caption{Our proposed circuit schematics for commercial MCU adaptation with scalable sensor arrays.}~\label{fig:circuit}
\end{figure*}

 Based on the system architecture mentioned above, we have conceptualized a schematic-level expandable system for general MCU integration (Fig. \ref{fig:circuit}). The CH-101 consists of 3 additional I/O pins on top of I2C communication lines for resetting, triggering, and readback operations. For data telemetry, a dedicated I2C level shifter (TCA9406) is used. Pull-up resistors are placed on each side of the bus for open-drain operation. Correspondingly, several bidirectional level shifters (TXB0104) are used to level shift the SPI bus and other peripheral I/O lines for 1.8 environment. Three SPI-based I/O expanders (MCP23S08) handle the control of 8 CH-101s. "RST" and  "PROG" pins are unidirectional, but "INT" pins are bidirectional with a non-industrial standard high-side open-drain drive, making them incompatible with the majority of the level shifters on the market. Therefore, additional buffers are added to convert the drive mechanism of the INT pin to the industrial-standard push-pull drives. Lastly, a power regulator (MIC5504) provides 1.8V power. \par

\end{document}